# Hanprome: Modified Hangeul for Expression of foreign language pronunciation

Wonchan Kim, Michelle Meehyun Kim

12/20/2024

**Abstract**

Hangeul was created as a phonetic alphabet and is known to have the best 1:1 correspondence between letters and pronunciation among existing alphabets. In this paper, we examine the possibility of modifying the basic form of Hangeul and using it as a kind of phonetic symbol. The core concept of this approach is to preserve the basic form of the alphabet, modifying only the shape of a stroke rather than the letter itself. To the best of our knowledge, no previous attempts in any language have been made to express pronunciations of an alphabet different from the original simply by changing the shape of the alphabet strokes, and this paper is probably the first attempt in this direction.

## 1 Introduction

This article focuses on the elements of the alphabet and their pronunciation. So let's take the word "element" as our first example. The pronunciation of the "e" varies depending on its position. “Dependent”, here too, the pronunciation of the “e” changes depending on position, and even when placed side by side with 'element', no rules can be found.

It's not just in English. In Russian, for example, vowel pronunciation changes, as shown in the following example: “Большой!”, “Отлично!”, “Хорошо!”

This variation in pronunciation can be confusing for learners. In modern times, a learner is not necessarily a human being but can also be an AI system. This kind of discrepancy in pronunciation and spelling can result in greater consumption of time and energy than necessary.[15]

While phonetic symbols such as IPA(international phonetic alphabet[14]) can be helpful for explanation, it is not practical to always rely on them. Is there a way to express these pronunciations using an everyday alphabet?

Learning the foreign language itself is, of course, an effective approach, but this is a privilege for a select few. Even so, it is impossible for them to learn all the languages they encounter. A more effective way would be to express the pronunciation of the foreign language using the alphabet used in everyday life.

One thing to note here is that when the word 'international' comes up, many people associate it with Esperanto, but Esperanto was created as a 'common language of equality for humanity' in opposition to English and French, which were powerful national languages at the time, being used as 'international languages'[22], and it was a movement that had nothing to do with the pronunciation itself. The number of consonants it can express is 23, and the number of vowel sounds it can pronounce is 5. [23]

From this perspective, we will adopt Hangeul -not as the Korean language, but as an alphabet- as an example of a possible candidate. We will examine its strengths and limitations and explore possible ways to improve it.

### 1.1 What is Hangeul?

Hangeul is a writing system created by King Sejong the Great in 1443 [1]. He created the 'phonetic alphabet' so that people could easily express their thoughts without having to learn thousands of Chinese characters ('ideographic characters'). In other words, Hangeul is inherently a ‘phonetic symbol’ from the beginning. At the time of its creation, Sejong created a form that could express not only Korean pronunciation but also Chinese pronunciation[2], hence the noble class considered reading and writing Chinese a great privilege and were very afraid of losing their acquired rights and protested very vigorously[3]. Hangeul, as a unique alphabet system, was designated a UNESCO World Heritage in 1997[4].

### 1.2 How Hangeul is written.

Since the details of reading and writing Hangeul are beyond the scope of this paper, only the key characteristics of Hangeul will be briefly explained here.
The basic components of Hangeul are as follows:

The consonants:

ㄱ ㄴ ㄷ ㄹ ㅁ ㅂ ㅅ ㅇ ㅈ ㅊ ㅋ ㅌ ㅍ ㅎ

The vowels:

ㅏ ㅑ ㅓ ㅕ ㅗ ㅛ ㅜ ㅠ ㅡ ㅣ

Simply put, in Hangeul, the elements of the alphabet are first combined horizontally or stacked vertically to form a syllable and then connected to form a word, while in English, the elements of the alphabet are arranged laterally and connected to form a word,

Consonants can be combined to create other consonants: ㄱ+ㄱ = ㄲ, ㄷ+ㄷ= ㄸ, ㅂ+ㅂ= ㅃ

Vowels can also be combined to create other vowels: ㅓ + ㅣ = ㅔ, ㅏ + ㅣ = ㅐ, ㅗ+ ㅣ = ㅚ.

The consonants and vowels of Hangeul combine to create a syllabary.

- First pronunciation vowel, ㅇ is placed as ‘no consonant’ symbol: 아 야 어 여

- Vowels with a long vertical line are written to the right of the consonant: 가 터 대 뻬

  Vowel with a long horizontal line is written below the consonant: 로 소 추 뚜

- The last consonant of a syllable is written under the vowel: 각 솔 쫌 밝

**1.3 How Hangeul is pronounced.**

As a basic element of letters, there is no difference between Hangeul and English alphabets. The corresponding pronunciations are organized in the chart below.

| consnt | pronunciation |
|---|---|
| ㄱ | g (gong) |
| ㄴ | n (nice) |
| ㄷ | d (door) |
| ㄹ | r (room) / l(cool) |
| ㅁ | m (moon) |
| ㅂ | b (bus) |
| ㅅ | s (sun) |
| ㅇ | silent / ng(song) |
| ㅈ | j (Japan) |
| ㅊ | ch (China) |
| ㅋ | k (Korea) |
| ㅌ | t (team) |
| ㅍ | p (pine) |
| ㅎ | h (home) |

| vowl | pronunciation |
|---|---|
| ㅏ | a (ice) |
| ㅑ | |
| ㅓ | (upon) |
| ㅕ | (young) |
| ㅗ | o (auction) |
| ㅛ | (yodel) |
| ㅜ | u (wood) |
| ㅠ | (use) |
| ㅣ | i (even) |
| | |
| ㅐ | (ant) |
| ㅔ | e (error) |
| | |
| ㅡ | (fr. jeu) |

**2 What are the strengths and limitations of Hangeul?**

Here, it absolutely does not mean that Hangeul is superior to any other alphabets, but rather to emphasize that Hangeul has the potential to be used as a phonetic symbol because it is an alphabet in which the letters and pronunciation correspond 1:1.

**2.1 The Vowels**

In Hangeul, no consonant or vowel is read differently on a case-by-case basis, and only the given pronunciation comes from a given letter.

For example, in the English word ‘dependent’, ‘e’ is read as three different sounds depending on its position, and in Hangeul, these different sounds are expressed as different vowels:

dependent 이, 에, 어

Here is another example.

and another ball game 애, 어, 오, 에이

The German umlaut sounds ä, ö, ü can also be simulated in the following way:

täglich / böse / dünn 애 / 외 / 위

The vowel ‘ㅡ’ can be very useful in expressing French pronunciation

un petit garçon (a little boy) 앙 쁘띠 갸르쏭

une petite maison (a little house) 윈느 쁘띠뜨 메종

The atmosphere that emerges from reading the following line in French cannot be conveyed by reading it in the English reading style, but it may be possible if read in Hangul.

"Elle est plus belle que les étoiles dans le ciel..." 엘 에 쁠뤼 벨르 끌레제뜨왈르 당 르 씨엘..

(She is more beautiful than the stars in the sky...)

The vowel ‘ㅢ’ enables the expression of the pronunciation of Russian ы, which is very important in Russian pronunciation.

Обычно ты быстро бежал.... 아븨치너 띄 븨스트러 비좔

(Usually you ran quickly....)

**2.2 The Consonants**

The 1:1 correspondence is true also for consonants.

Paris is written ‘**Paris**’ in German, English, Portuguese, and Spanish,

but ‘**Parigi**’ in Italian, ‘**Παρίσι**’ in Greek, and ‘**Париж**’ in Russian.

(The Trojan shepherd with the Golden Apple was **Πάρις**,

and the Russian word for ‘betting’ is happens to be **Пари**.)

Although the spellings look similar, the pronunciation is different in each language.

The Hangeul expressions below clearly show these fine differences in pronunciation

French: 빠히, German: 파리스, English: 패리스, Spanish: 파리스, Italian: 빠리지,

Russian: 빠리쥐, Greek: 빠리시, Portuguese: 빠리(시), the Trojan shepherd: 빠리(스)

It would be a difficult task to explain the pronunciation differences in each language using the English alphabet, but as can be seen below, the pronunciations can be expressed 1:1 in Hangeul. (Note for those who cannot read Hangul yet: If the letter shapes are different, they represent different sounds.)

Not only this p in the examples above, t and c in these languages show the same ‘dense (linguistics terminology: ‘fortis’) characteristics. Perhaps it is rather better to say that they are present in French, Spanish, Italian, Russian, and in other languages, but not in English.

French: pourquoi, tableau, cadeau

Italian: porta, tavolo, casa

Spanish: porque, también, cuando

Russian: привет, также, когда.

Hangeul can express these sounds with ‘double consonants’ : ㄲ, ㄸ, ㅃ

porta 뽀르따 vs. peace 피스

tableau 따블로 vs. table 테이블

cadeau 꺄도 vs. card 카드

Written in Hangeul, the difference between plosives and fortis is distinguishable.

pizza deliziosa” (delicious pizza) 삐짜 델리치오자

Le Petit Prince (The Little Prince) 르 쁘띠 프랑스

**2.3 What are the limitations of Hangeul?**

However, Hangeul has its limitations.

For example, several pronunciations in the English sentence below cannot be expressed.

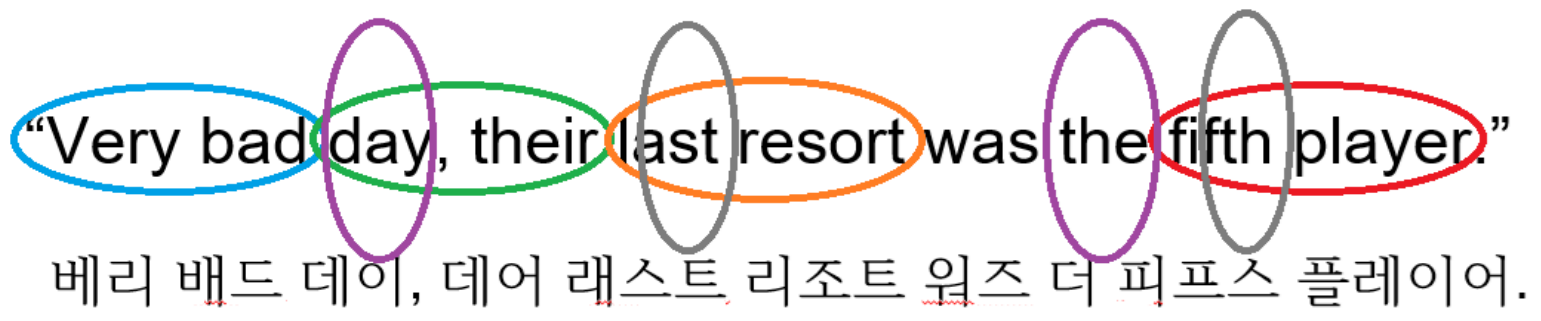

This reveals some difficulties related to the expression of English pronunciation.

The following consonant pairs are expressed with the same alphabet letter.

‘very/bad as ‘ㅂ’, day/their as ‘ㄷ’, last/resort as ‘ㄹ’, last/fifth as ‘ㅅ’, fifth/player as ‘ㅍ’.

If only these few problems presented here can be eliminated, there is a possibility of replacing phonetic symbols with "natural language" Hangeul instead of using the ‘unnatural, invented’ alphabets[11] or the unfamiliar symbols like θ, ð [12] to express the pronunciation of a foreign language. The "phonetic symbols" mentioned here refer to the level of tools that ordinary people can use in their daily lives, and are not of the same character as the IPA symbols that apply to all languages, but are applied in different sets depending on the language treated.

## 3 Modification of the Hangeul Alphabet

It is hard to imagine changing Hangeul itself to express the pronunciation of foreign languages. It may be like an attempt to add the German umlaut or the French spelling variations or even Cyrillic-like new letters to the English alphabet. Then, how about changing the shape of the alphabets? In fact, various shapes of the alphabet already exist, which are called fonts. There is no way that adding one more here would make it more special.

### 3.1 Any possibility to overcome these limitations?

To explore some possibilities, let's look at the following word, which means 'Valley'.

밸리

Here, we need to somehow change the shape of the letter ㅂ, which signifies the b not v sound so that it becomes a new shape that signifies the v sound.

But, with a constraint that the shape of the letter not to be changed, this is a kind of paradox.

But if carefully thought, the pronunciation does not come from the shape of the letter. In Spanish, both v and b are pronounced as b. Furthermore, Russians even write в and read it as v. (б is pronounced as b.)

How about bending the first stroke? That will not affect the basic shape itself.

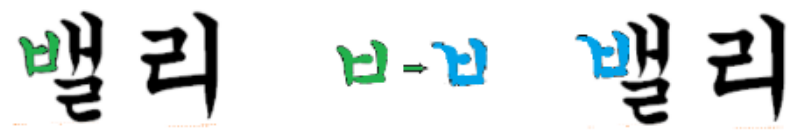

However, this doesn't seem like a plausible solution. Since the shape of the strokes can be changed according to the taste of the writer, making it impossible to know whether the intention related to the distorted letters is simply an attempt to satisfy the writer’s whim or a deformation for a specific purpose, as desired here.

A little deeper thought leads us to another possibility.

What if the stroke were to swell into a two-dimensional shape with area?

Of course, the shape of the line after swelling can vary, and it is true that there are still concerns about the reproducibility of this shape. But if we think about it carefully, there may be no reason to worry. Because, nowadays, people use more computer-generated fonts rather than hand-written ones. Thus, the reproducibility of fonts reproduced in this way is guaranteed.

The shape of the stroke can now be changed either by twisting or by swelling the line, as following:

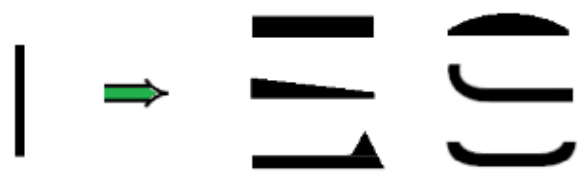

### 3.2 Target stroke finder

**Consonants**

Now the important task in transforming the alphabet is to decide which stroke of the alphabet to transform. The stroke selected should be the most visible one among the letters, and the task to find it is as follows.

Due to the compositional characteristics of Hangeul letters, consonants are located at the top of the letter and strokes start from the top right. So the search starts at the top left corner of the frame covering the text. The pixel cluster that contains the first pixel encountered at this time is the target consonant.

In that group, target strokes are determined in the following order of priority:

- strokes moving toward the right,
- strokes moving downward,
- strokes moving diagonally

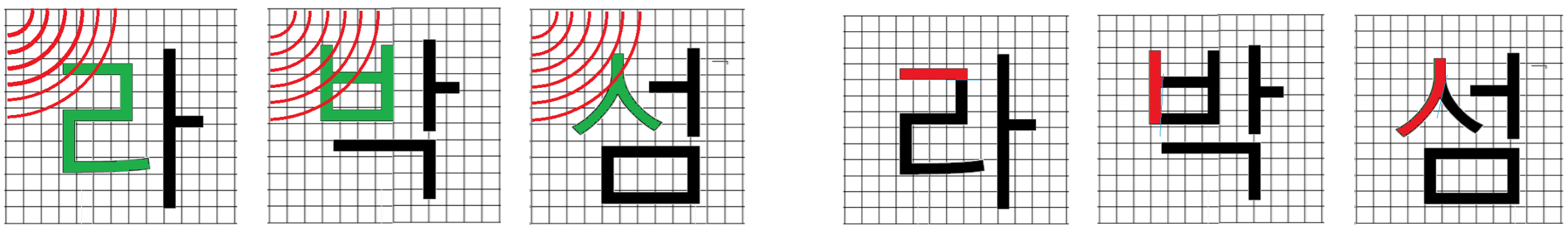

**Vowels**

Due to the compositional characteristics of Hangeul, vowels contain the longest stroke in the letter shape. Therefore, it is a priority to find the most constituent elements in the horizontal or vertical direction among the target letters. If the longest line is selected from this pixel collection found this way, it is the target stroke.

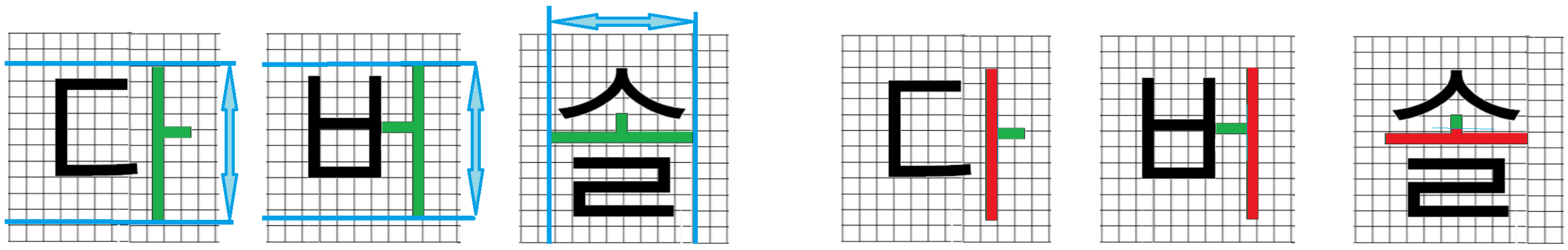

### 3.3 Target stroke modifier

The simplest way is to uniformly increase the thickness of this selected stroke as shown in the figure below.

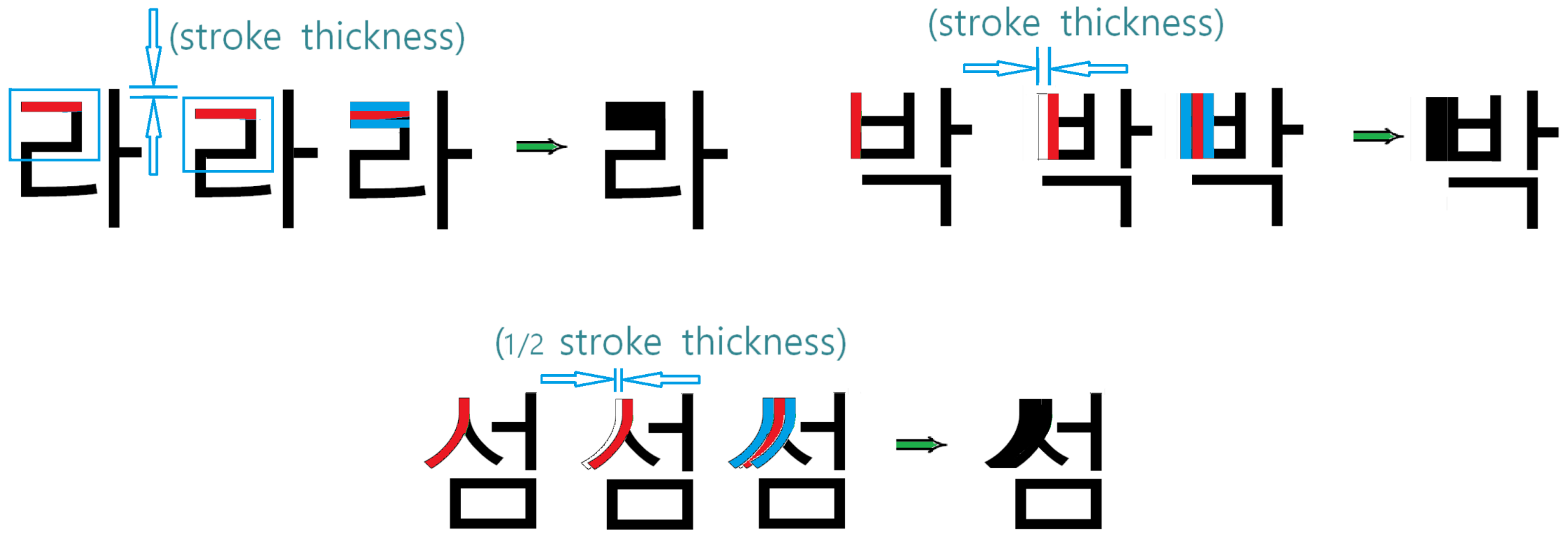

Of course, this shape can be modified further through geometric operations.

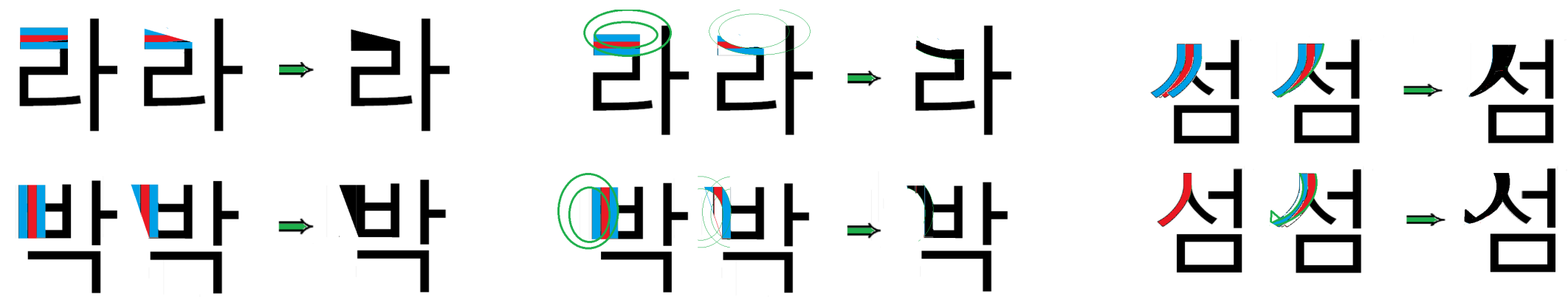

The figure below also shows the modified vowels using this method.

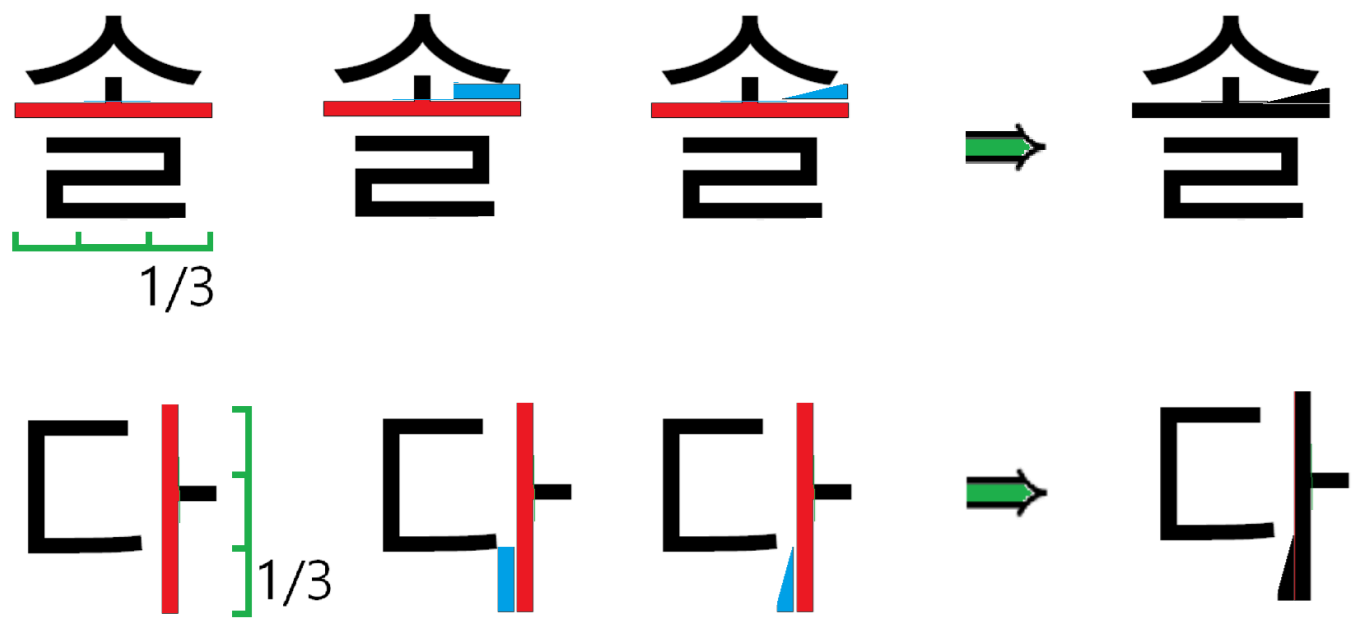

## 4 The Modified Hangeul

### 4.1 Modified Consonants

In order to express the English pronunciation mentioned above, the consonant part of Hangeul is now transformed as follows:

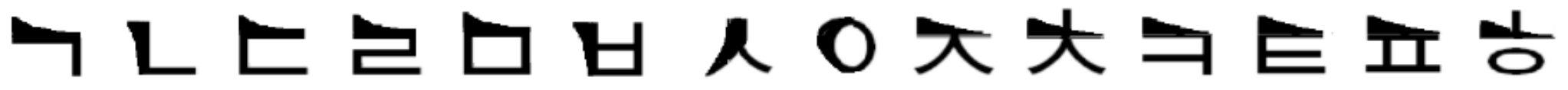

Using this modified consonant shape, we can distinguish the pronunciations of the five cases that were problematic above.

Very bad day, their last resort was the fifth player.
베리 배드 데이, 데어 래스트 리조트 워즈 더 피프스 플레이어.

| | |
|---|---|
| pace / face | 페이스 / 페이스 |
| best / vest | 베스트 / 베스트 |
| sink / think | 싱크 / 싱크 |
| day / they | 데이 / 데이 |
| row / low | 로우 / 로우 |

Also, the modified Korean consonant ㅋ can be used to express the harsh and strong h/k sound ch/x/خ, which is not found in English but is frequently found in German[6]/Russian/Arabic. Even in Spanish, where h is silent, the j sound is sometimes nearly as strong as this.

| | |
|---|---|
| Bach | 바카 |
| хорошо | 카라쇼 |
| خالد | 칼리드 |

Also, we know already that the pronunciation of r in French[8]/Portuguese is different from English, but if we still want to note that explicitly, we can use the modified consonant corresponding to ㅎ for that purpose.

| | | |
|---|---|---|
| Real | 히알 (Portugal) | 헤알 (Brazil) |
| Ronaldo | 후날두(Portugal) | 호날두(Brazil) |

The point that needs to be emphasized here again is that this is not a simple font operation, but a transformation that induces different pronunciations and different meanings.

### 4.2 Modified Vowels

Now that we have solved the problem in the consonant part, it is time to look for improvement in the vowel part.

Since the basic syllable composition of Hangeul is a ‘combination of consonants and vowels’, it is not possible to express a silent sound in Hangeul. For example, the word ‘first’ is originally a one-syllable word, but if it is written as ‘퍼스트’ in Hangeul, it becomes a three-syllable word. Not only the English "st", but also, for example, the r sound in Goethe's "Die Leiden des jungen Werther" or the French name Sartre is still written monosyllabically in Hangul. To resolve this inconvenience, a ‘absence of vowel’ short — is introduced, in correspondence to the 'absence of consonant' ㅇ.

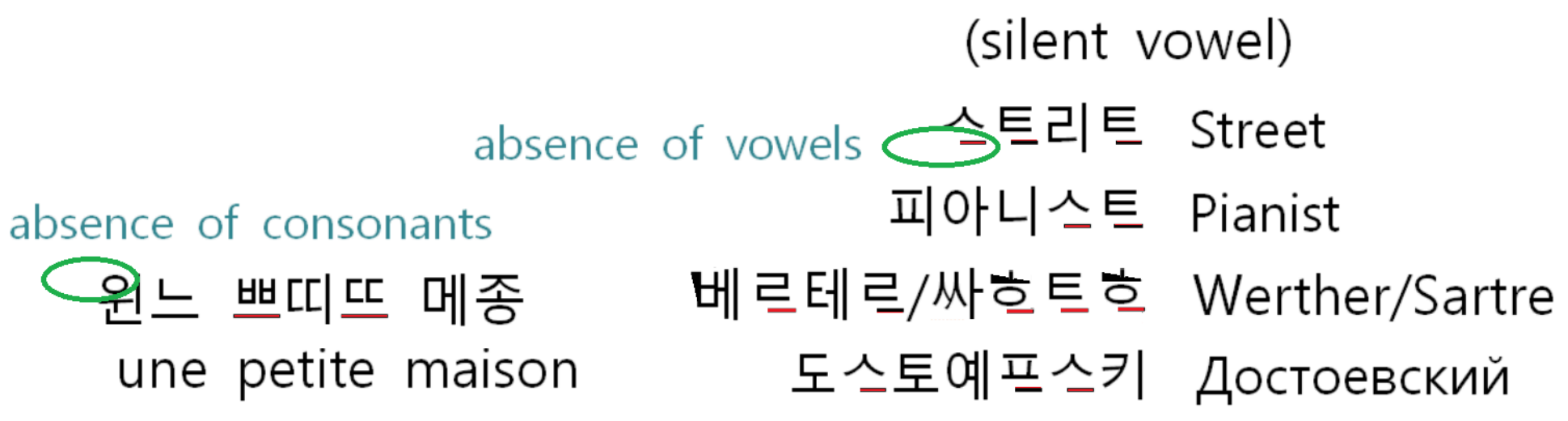

Also, in English, ‘vowel + r’ is used as a kind of ‘unit pronunciation’[5], and a method of expressing this in Hangeul is presented.

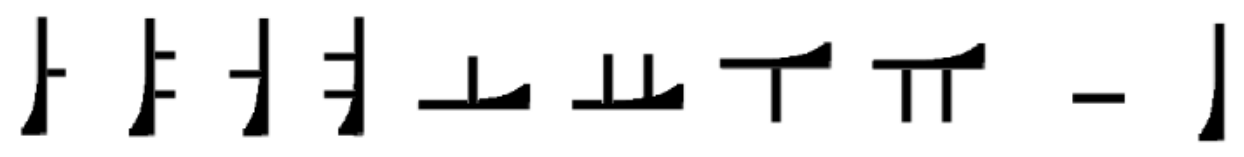

Now, if we apply this modified vowel to the example given in the previous section on consonants, it becomes as follows.

Very bad day, their last resort was the fifth player.

베리 배드 데이, 데어 래스트 리조트 워즈 더 피프스 플레이어.

By using these modified vowels, it is now possible to express typical American pronunciation with Hangeul.

| | |
|---|---|
| 발코니 / 바코드 | balcony / barcode |
| 얄타 / 야드 | Ялта / yard |
| 다이얼 / 다이어 | dial / dire |
| 월드 / 워드 | world / word |
| 소셜 / 요크셔 | social / Yorkshire |
| 코드 / 콜드 / 코드 | code / cold / cord |
| 욜로 / 뉴욕 | yolo / New York |

## 4.3 Modified Hangeul Alphabets

The tables below show the original form of Hangeul and its modified forms.

Consonants

| org. | pronunciation | mod. | pronunciation |
|---|---|---|---|
| ㄱ | g (gong) | | |
| ㄴ | n (nice) | | |
| ㄷ | d (door) | ㄷ | th (that) |
| ㄹ | r (room) | ㄹ | l (land) |
| ㅁ | m (moon) | | |
| ㅂ | b (bus) | ㅂ | v (volume) |
| ㅅ | s (sun) | ㅅ | th (three) |
| ㅇ | silent / ng(song) | ㅇ | ng (fr. long) |
| ㅈ | j (Japan) | | |
| ㅊ | ch (China) | | |
| ㅋ | k (Korea) | ㅋ | ch(ger)(Bach) |
| ㅌ | t (team) | | |
| ㅍ | p (pine) | ㅍ | f (fine) |
| ㅎ | h (home) | ㅎ | r (fr) (arbre) |

Vowels

| org. | pronunciation | mod. | pronunciation |
|---|---|---|---|
| ㅏ | a (ice) | ㅏ | (art) |
| ㅑ | | ㅑ | (yard) |
| ㅓ | (upon) | ㅓ | (earth) |
| ㅕ | (young) | ㅕ | (yearn) |
| ㅗ | o (auction) | ㅗ | (organ) |
| ㅛ | (yodel) | ㅛ | (Yorkshire) |
| ㅜ | u (wood) | ㅜ | (fr. amour) |
| ㅠ | (use) | ㅠ | |
| ㅣ | i (even) | ㅣ | (fr. dire) |
| | | | |
| ㅐ | (ant) | ㅐ | (ger. Ärger) |
| ㅔ | e (error) | ㅔ | (ger. Erde) |
| | | | |
| ㅡ | (fr. jeu) | ㅡ | silent |

## 4.4 From the user's perspective

No matter how appealing this idea of modified alphabet may be, it would be impractical if it were inconvenient to input to devices. Fortunately, however, in the case of Hangeul there is a unique possibility. Unlike English, Hangeul does not distinguish between uppercase and lowercase letters, so as can be seen in the smartphone input keyboard below, when the Shift key is pressed for uppercase letters of English, there is a lot of unused space. The alphabet variants we want to use can be placed on our keyboard in these unused spaces. Since not all the Hangeul variants are needed, there is plenty of space to place them.

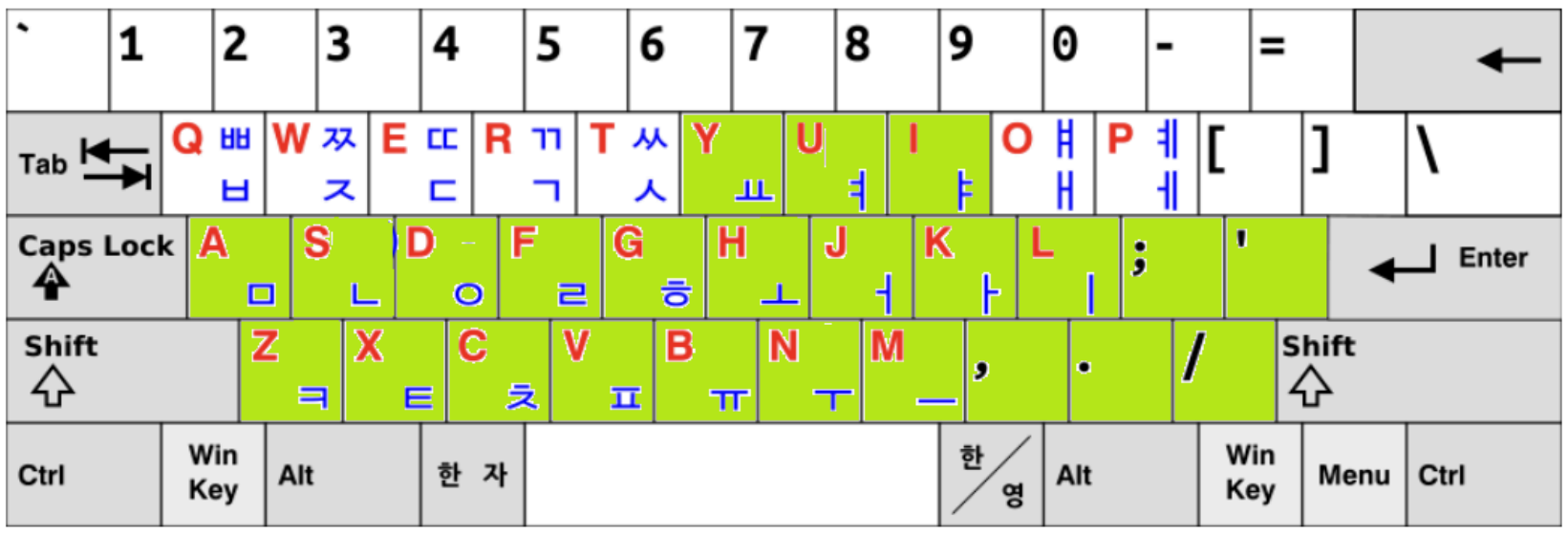

Therefore, hardware manufacturers do not need to change the keyboard structure and users can also use the keyboard they are already familiar with. All that is needed here is to display the symbol for the corresponding modified alphabet letters on the keyboard at that location. The subsequent OS-related work is omitted here as it does not fit the nature of this paper.

## 5 Application of modified Hangeul Alphabets

### 5.1 English

Below is an example of applying this modified Hangeul alphabet to English.

Listen to the voice for bread and peace!
리쓴 투 더 보이스 퍼 브레드 앤 피스

It is up to the user's preference how to modify the shape of the strokes.

The wars are making the world worse.
더 워즈 아 메이킹 더 월드 워스.

There's a fire starting in my heart.
데어즈 어 파이어 스타팅 인 마이 하트.

Sartre loved her life there through thick and thin.
사르트르 러브드 허 라이프 데어 스루 시크 앤 신.

### 5.2 Application to Italian and Spanish,

In Italian, which developed directly from Latin, spelling and pronunciation are relatively consistent. This characteristic is also maintained to some extent in Spanish. However, this does not mean that there are no irregularities.

Perhaps the most useful application of this modified Hangeul is in these languages.

First, pronunciation differences that cannot be expressed by reading them in the English manner can be expressed in Hangeul.

Furthermore, regardless of whether a grammatical rule is applied or whether the pronunciation is pronounced in its European country of origin or in a South American country, as in the case of Spanish or Portuguese, regardless of the context, the resulting pronunciation can be expressed differently in Hangeul.

Unlike in Spanish, in Italian the letter s changes to the sound z when placed between vowels.

(sp) casa 까싸 (it) casa 까자

(sp) desierto 떼씨에르또 (it) deserto 데제르또

The sound of the letter z in Italian varies depending on grammatical rules.
In Hangeul, these differences can be expressed by distinguishing them.

La ragazza parla con suo zio delle zebre.
라 라가짜 빠를라 꼰 수오 치오 델레 제브레.

The girl talks to her uncle about zebras

On the other hand, in Spanish, this z is pronounced like the English s.
The irony is that in most parts of Spain this z is pronounced like the English th.
Not only z, the cielo, cerveza, cita (c before e/i) are pronounced as English th.
In all other Spanish-speaking areas, it is simply pronounced as s.
So, to clearly express the pronunciation of s in Spanish, c is employed.
Thus the Spanish equivalent of ‘zero’ is ‘cero,’ and for ‘zebra’ it's ‘cebra.’

People can easily get confused when they hear the explanations like the above.
Regardless of the grammar, Hangeul can express the resulting pronunciation as it is.
Below are the pronunciation differences between Spanish and Italian sentences.

(sp) El sol brilla en el cielo azul.
엘 쏠 브리야 엔 엘 씨엘로 아술
(it) Il sole splende nel cielo azzurro.
일 솔레 스플렌데 넬 치엘로 아쭈로

The sun is shining in the blue sky.

The following two sentences in Spanish and Italian of same meaning with similarity in spelling, sound quite different due to the difference in pronunciation of c and z.

(sp) El cine está cerca de la plaza principal.
엘 씨네 에스따 쎄르까 델라 쁠라사 쁘린씨빨
(it) Il cinema è vicino alla piazza principale.
일 치네마 에 비치노 알라 삐아짜 쁘린치빨레.

The cinema is near the main square.

**5.3 Application to German**

There is an important pronunciation of ch, which is not present in English.
The harsh and strong h sound [7] (or perhaps closer to the k sound) that frequently appears in German can now be matched with the modified Hangeul.
Let's start this pronunciation example with a sentence from Nietzsche.[13]

Was mich nicht umbringt, macht mich stärker.
바스 미키 니키트 움브링트 마카트 미키 슈태르커
What does not kill me makes me stronger.

Menschenrechte? So etwas gibt es ja doch nicht.
멘션레키테 조 에트바스 깁트 에스 야 도코 니키트
Human rights? There is no such thing.

Dort ist es ja wirklich fürchterlich!
도트 이스트 에스 야 비르크리키 퓌키터리키!
It's really terrible there!

The first thing that catches your eye when you come across German sentences is the umlaut.

If the original mouth shape is kept during pronouncing the compound vowels in Hangeul, they will produce almost the same sound.

Es ist ärgerlich, dass es überhaupt keine Lösung gibt.
에스 이스트 애거리키 다스 에스 위버하웁트 카이네 뢰중 깁트
It is annoying that there is no solution at all.

Er spricht über die schönen Länder.
에어 슈프리키트 위버 디 쇠넨 랜더.
He talks about the beautiful countries.

In addition to this ‘ch’ and Umlaut, there are a few more differences in German pronunciation to note: the s right before a vowel is pronounced as z, while the z itself is pronounced as ‘is’. Also, w is pronounced as v, and g before consonants is pronounced as ch. And, voiced sounds such as b and d at the end of words end in voiceless sounds such as p and t.

All of these points to note are included in the sentence below, and there is also no problem in displaying these pronunciations in Hangeul.

Dazu benötigst du Zwiebeln und Salz.
다추 베뇌티키스트 두 츠비벨른 운트 잘츠.
For this you need onions and salt.

### 5.4 Application to Russian

Russian is a language so different from English in its pronunciation system [19] that it is advised that gymnastics of the vocal organs is necessary[9]. But In Russian language itself, the pronunciation and letters are relatively consistent, but there is some flexibility.

The vowel sounds vary depending on the presence or absence of accents and their placement.[10]

Преступление и наказание Достоевского
쁘레스뚜쁠리니예 이 나까자니예 도스따예프스카바
Crime and Punishment by Dostoevsky

The fortis sound appears frequently in Russian as well as in Spanish, Italian and French. Regardless of the cases, the resulting pronunciation can be expressed in Hangeul.

Каждое утро, я пью кофе и читаю книгу.
까쥐다에 우뜨러 야 삐유 코피예 이 치따유 크니구.
Every morning, I drink coffee and read a book.

What is unique about Russian is that in the case of ти, these т pronunciation changs, becoming closer to the Italian zz pronunciation. Similarly the д followed и changes its pronunciation closer to English z pronunciation. Whatever the case, the resulting pronunciations can also be expressed in Hangeul.

Поля купила красивую картину.
빨랴 꾸삘라 크라시부유 까르찌누
Polya bought a beautiful painting.

Наш директор Диана сидит на диете.
나쉬 지렉터르 지아나 씨짓 나 지에떼
Our director Diana is on a diet.

**The** sentences in the following contain the Russian-specific ы pronunciation, which can only be expressed with Hangeul vowels.

Внутри комнаты было тихо и уютно.
브누트리 꼼나띄 뷜러 치코 이 유트너
Inside the room, it was quiet and cozy.

Тихий парк полон красивых цветов.
찌키이 빠르크 뽈론 크라씨븨크 츠비토프
The quiet park is full of beautiful flowers.

Now armed with a powerful tool called modified Hangeul, we can a little bit more faithfully reproduce the original Russian pronunciation.

Мир хлеба и фруктов, львов и волков.
미 흘레바 이 프룩토프, 류보프 이 볼코프.
the world of bread and fruits, lions and wolves.

### 5.5 Application to French

Perhaps the first thing that foreigners feel when they hear French is the feeling that there are a lot of ‘uvular r’ sounds, and the pronunciation of r can be said to have traces of r in the pronunciation of h.

Pierre Cardin 삐에흐 꺄흐당
tous les jours 뚤레쥬흐 every day
au cas par cas 오꺄빠흐꺄 case by case
l'ombre des arbres 롱브흐 데 자흐브흐 The shade of the trees

In French, the ng pronunciation ends as if the pronunciation is not yet tied up.
So, in this modified Hangeul, this nasal ng is written as ㅇ with the bottom part cut off.

Il pense souvent à son cousin à la campagne.
일 빵스 쑤벙 아 쏭 꾸장 아 라 깡빠니으.
He often thinks of his cousin in the countryside.

French is characterized by many words ending in vowels.[8]
Unlike in English, where the e at the end of a word is silent (nice, same, robe),
in French, the letter e has a sound. homme: 옴므 toute: 뚜뜨 belle: 벨르
Even pronunciations ending in -lle, bouteille: 부떼이으 Marseille: 마흐세이으
In the following cases, similar sound. peu: 쁘 bleu: 블르 beurre: 버흐

The typical dense pronunciation of the French p also changes to the English p pronunciation when combined with r, and the difference in pronunciation can also be expressed in Korean.

C'est trop tôt. / C'est trop tard.
쎄 트로 또 쎄 트로 따흐 It's too early. / It's too late.

Notamment, Notre Dame.
노따망 노트르 담므 In particular, Notre Dame

Below are some pronunciations that would not be easy to express in other languages.

la petite maison sur la colline
라 쁘띠뜨 메종 쒸 라 콜린느 the little house on the hill

Oui, il aime les oiseaux et les poissons.
위 일 엠므 레좌조 에 레 빠쏭 Yes, he likes birds and fish.

There is no need to explain which is closer to the original pronunciation when reading the following Hangeul expression or the English-alphabet expression in English.

L'homme a voulu monter vers les étoiles.
롬마 불뤼 몽테 베르 레제뜨왈르.
Humans want to climb towards the stars.

Même les très grandes étoiles nous paraissent petites.
멤므 레 트레 그랑데제뜨왈르 누 빠레스 쁘띠뜨
Even very big stars appear small to us.

**5.6 Application to Chinese**

Chinese and Korean are completely different languages from a grammatical perspective [24]. For example, in Chinese, verbs come before the object, but in Korean, verbs come after the object. Also, while Korean can express various tenses through conjugation of endings, in Chinese, such expressions are severely limited. However, from a word perspective, since they are languages belonging to the same cultural sphere, there are many similar pronunciations, just as French and English have many words with similar spellings.

While Hangeul is a phonetic alphabet, Chinese characters are ideographic characters that represent meanings[18]. Therefore, when people learn Chinese, they have to learn the pronunciation separately. In Taiwan, they use Zhùyīn fúhào(注音符號[16]), and in China, they use Hanyu Pinyin (汉语拼音[17]), which represents the Chinese pronunciation in alphabet form.

However, if you look at the nature of this 汉语拼音, it feels more like "This pronunciation should be written like this so that it doesn't overlap with other pronunciations..." rather than meaning "Read the Chinese characters like this in English..." However, due to the fundamental pronunciation differences between Chinese and English, this can only function as a 'hint', and there are many differences in actual pronunciation.

For example, thank you in Chinese 謝謝 is written as xièxiè in Hanyu Pinyin, but how many foreigners would read it as 'sye,sye'? This is the same for vowel pronunciation as well as consonants. The Hanyu Pinyin for the Chinese currency unit '元' is yuán, which is closer to the actual pronunciation in German üen. Also, beer is written as 啤酒 píjiǔ, but in reality, it is a continuous pronunciation of 'iou' rather than a simple 'í'.

Modified Hangeul can also express Chinese pronunciation almost exactly. Below, we have arranged the modified Hangul letters based on the 汉语拼音 consonant chart in Wikipedia. The important point is that these modified Hangeul ‘꽃한글’ letters correspond 1:1 with the 汉语拼音.

(Although the contents of vowels will not be discussed in this paper, the vowels pronounced with 汉语拼音 can be evaluated much further from the original Chinese pronunciation than the consonant part, and of course, it can be organized much more naturally as ‘꽃한글’.)

In order to match the consonants of Chinese characters and modified Hangul consonants on a 1:1 basis, the following measures were taken:

- In Chinese, there is no 'r' pronunciation and the default pronunciation is 'l', as in 老人(lǎorén) 理想(lǐxiǎng), the basic Hangeul 'ㄹ' is matched to this 'l'.

- Instead, modified Hangeul is placed in the positions of the 卷舌音(scroll tongue sounds zh, ch, sh, and r ) that foreigners have the most difficulty pronouncing.

- As in 词典(cídiǎn) 次要(cìyào), the ci always produces a pronunciation similar to '츠', while as in 氣流(qìliú), 其他(qítā), the qi always produces a pronunciation similar to '치', so this is also matched to the unmodified Korean letter '츠' and ‘치’.

| 漢語拼音 | Hangeul | in 漢語拼音 | in Hangeul |
|---|---|---|---|
| b | ㅂ | 北京 Běijīng | 뻬이징 |
| p | ㅍ | 啤酒 Píjiǔ | 피지오우 |
| m | ㅁ | 母愛 Mǔ'ài | 무아이 |
| f | ㅍ | 方法 Fāngfǎ | 팡파 |
| d | ㄷ | 東西 Dōngxī | 똥씨 |
| t | ㅌ | 統一 Tǒngyī | 통이 |
| n | ㄴ | 你好 Nǐ hǎo | 니하오 |
| l | ㄹ | 老師 Lǎoshī | 라오시으 |
| g | ㄱ | 個人 Gèrén | 꺼런 |
| k | ㅋ | 康熙 Kāngxī | 캉시 |
| h | ㅎ | 皇帝 Huángdì | 황띠 |
| j | ㅈ | 進行 Jìnxíng | 찐싱 |
| q | 치 | 青島 Qīngdǎo | 칭따오 |
| x | ㅅ | 喜歡 Xǐhuān | 씨환 |
| zh | ㅈ | 中國 Zhōngguó | 쫑구오 |
| ch | ㅊ | 常規 Chángguī | 창궤이 |
| sh | ㅅ | 上海 Shànghǎi | 샹하이 |
| r | ㄹ | 日本 Rìběn | 르번 |
| z | ㅉ | 自由 Zìyóu | 쯔요우 |
| c | 츠 | 刺客 Cìkè | 츠커 |
| s | ㅆ | 四川 Sìchuān | 쓰촨 |

First, for comparison, here is a Chinese proverb: Confucius's words from 2500 years ago [20].

四海之內皆兄弟也。Sìhǎi zhī nèi jiē xiōngdì yě.

씨하이 즈네이 지에 숑디 예 "We are all brothers within the four seas."

Here, the modified Hangul is used only in one place, the pronunciation of 之.

Another proverb from the same book.

先做学生后做先生。Xiān zuò xuéshēng hòu zuò xiānshēng.

셴 쭈오 쉬에셩 호우 쭈오 셴셩 "Be a student first and then a teacher. "

Here, the only part that requires modified Hangul notation is the pronunciation of 生.

Another passage from the Analects of Confucius[21].

三思而后行。Sānsī ér hòu xíng

싼씨 아ㄹ 호우 씽 "Think thrice before you act"

Here, the only part that requires modified Hangul notation is the pronunciation of 而.

In the three examples above, the reason for mentioning that only one pronunciation is to be displayed in modified Hangul was to emphasize that most Chinese pronunciations can be expressed in plain Hangeul.

Now, an example to emphasize the pronunciation of scrolled tongue sounds (卷舌音).

十年树木，百年树人。 Shínián shù mù, bǎinián shù rén

스녠 슈무 바이녠 슈런

"10 years to cultivate a wood, 100 years to cultivate a man"

In Chinese, tones are just as important as pronunciation. These tones are part of Chinese pronunciation. As can be seen in the figure below, the meaning changes completely depending on the tones. [25], [26]

mā má mǎ mà ma

媽 麻 馬 罵 嗎

In fact, since most pronunciations in Chinese can be nearly faithfully expressed in Hangeul, this modified Hangeul is more useful for notating tones, as shown in some examples below.

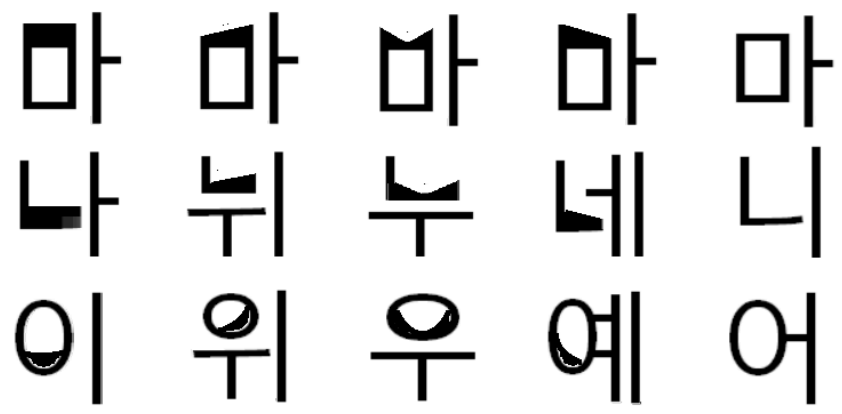

Below are some examples of sentences where Chinese pronunciation and tones are expressed with a mixture of modified Hangeul.

你在干什么?

Nǐ zài gàn shénme?

니 짜이 깐 션머 — What are you doing?

我的本意不是那样的。

wǒ de běnyì búshì nàyàng de.

워 더 번이 부-시 나양 더 — That wasn't my intention.

不知道有多高兴。

Bù zhīdào yǒu duō gāoxìng.

뿌 쯔따오 요우 뚜오 까오씽 — I don't know how happy I am.

**6. Conclusion**

This paper proposed a new method for expressing foreign language pronunciation by converting the stroke shape of Hangeul from a one-dimensional concept to a two-dimensional concept. By showing examples of applying this modified Hangeul, which provides a one-to-one correlation between letters and pronunciation, to English, German, French, and Russian, it is shown that users can employ this modified Hangeul as a kind of pronunciation symbol for foreign languages, maintaining the keyboard layout of the input device and the related software system currently being used.

**Authors**

Wonchan Kim, Professor Emeritus, Seoul National University, wonchank@snu.ac.kr

Michelle Meehyun Kim, Professor of Economics, Mission College, Santa Clara